%% file: main.tex
\documentclass[11pt,a4paper]{article}
\usepackage[hyperref]{acl2021}
\usepackage{times}
\usepackage{latexsym}

\usepackage{enumitem} 

\usepackage{microtype}

\usepackage{url}

\usepackage{tabu} 
\usepackage{bbm}
\usepackage{bm}
\usepackage{transparent}
\usepackage{CJKutf8}
\usepackage{nccmath}
\usepackage{xcolor}

\usepackage{mathtools}
\usepackage{balance}

\usepackage{url}

\usepackage{tikz}
\usepackage{tikz-qtree}

\usepackage{graphicx}			

\usepackage{mathpartir}
\usepackage{pifont}
\usepackage{booktabs}
\usepackage{tabularx}
\usepackage{makecell}
\usepackage{multirow}
\usepackage{array}

\usepackage{diagbox}

\usepackage{proof}
\usepackage{xspace}
\usepackage{multirow}
\usepackage{forest}
\usepackage{subcaption}
\usepackage{float}
\usepackage{dsfont}
\usepackage{amsmath}
\usepackage{amssymb} 
\usepackage{amsthm}
\newcommand{\namecite}[1]{\newcite{#1}}
\input{defs}
\definecolor{chocolate}{rgb}{0.28, 0.02, 0.03}
\definecolor{PaleGreen}{rgb}{0.33, 0.545,0.33}
\definecolor{colorC0}{RGB}{51,113, 169}
\definecolor{colorC1}{RGB}{243,130,37}
\usepackage{array}
\usepackage{diagbox}
\usepackage{algorithm}
\usepackage[noend]{algpseudocode}
\aclfinalcopy 

\DeclarePairedDelimiter\ceil{\lceil}{\rceil}

\title{Direct Simultaneous Speech-to-Text Translation \\Assisted by Synchronized Streaming ASR
\thanks{\ See our translation examples and demos at 
\href{https://littlechencc.github.io/SimulST-demo/simulST-demo.html}{\scriptsize https://littlechencc.github.io/SimulST-demo/simulST-demo.html}.
}
}

\author{  Junkun Chen $^{1}$ \quad
Mingbo Ma $^{2}$ \quad
Renjie Zheng $^{2}$ \quad
Liang Huang $^{1,2}$
\\
$^{1}$Oregon State University, Corvallis, OR, USA \\
$^{2}$Baidu Research, Sunnyvale, CA, USA \\
\texttt{chenjun2@oregonstate.edu, mingboma@baidu.com} \\}

\date{}

\begin{document}
\begin{CJK}{UTF8}{gbsn}

\maketitle
\begin{abstract} 
Simultaneous speech-to-text translation is widely useful in many 
scenarios. 
The conventional cascaded approach uses a pipeline of streaming ASR followed by simultaneous MT, 
but suffers from error propagation and extra latency. 
To alleviate these issues, recent efforts 
attempt to directly translate the source speech into target text simultaneously,
but this is much harder  due to the combination of two separate tasks.
We instead propose a new paradigm with the advantages of both cascaded and end-to-end approaches.
The key idea is to use two separate, but synchronized, decoders on streaming ASR and direct speech-to-text translation (ST), respectively,
and the intermediate results of ASR guide the decoding policy of (but is not fed as input to) ST.
During training time, we use multitask learning to jointly learn these two tasks with a shared encoder.
En-to-De and En-to-Es experiments on the MuST-C dataset demonstrate that our proposed technique achieves substantially better translation quality at similar levels of latency.

\end{abstract}

\section{Introduction}
\input{intro}

\section{Preliminaries}
\input{prelim}

\section{Direct Simultaneous Translation with Synchronized Streaming ASR}
\input{methods}

\section{Experiments}
\input{exp}

\vspace{-0.1cm}
\section{Conclusion}
\vspace{-0.2cm}

We proposed a simple but effective ASR-assisted simultaneous E2E-ST
framework. 
The streaming ASR module can
guide (but not give direct input to) the \waitk policy for simultaneous translation.
Our method improves ST accuracy  with similar 
latency.

\section*{Acknowledgments}
This work is supported in part by NSF IIS-1817231 and IIS-2009071.

\bibliographystyle{acl_natbib}
\bibliography{main}

\end{CJK}
\end{document}

%% file: defs.tex
\newcommand{\tuple}[1]{\ensuremath{\langle {#1} \rangle}}

\newcommand{\notes}[1]{}

\theoremstyle{definition}

\theoremstyle{plain}

\newcommand{\vecz}{\ensuremath{\mathbf{z}}}

\newcommand{\vecs}{\ensuremath{\mathbf{s}}\xspace}

\newcommand{\ith}[1]{\ensuremath{i^{{th}}}}

\newcount\permx
\newcount\permy
\def\permdot#1#2{
\permx=#1 \advance\permx by-1
\permy=#2 \advance\permy by-1
\psframe[fillcolor=black, fillstyle=solid]
(\permx,\permy)(#1, #2)
}

\newcommand{\argmax}{\operatornamewithlimits{\mathbf{argmax}}}

\newcommand{\toptop}{\operatornamewithlimits{\mathbf{top}}}

\newcommand{\startsym}{\mbox{\scriptsize \texttt{<s>}}\xspace}

\newcommand{\boxnum}[1]{{\setlength{\fboxsep}{1pt}\raisebox{1pt}{\hspace{1pt}\fbox{\tiny #1}\hspace{1pt}}}}
\newcommand{\ind}[1]{\ensuremath{_{\kern-0.5pt\boxnum{#1}}}}

\newcommand{\vecx}{\ensuremath{{\mathbf{x}}}\xspace}
\newcommand{\vecy}{\ensuremath{{\mathbf{y}}}\xspace}

\def\namecite{\newcite}

\newcommand{\smallnt}[1]{\ensuremath{_{\mbox{\tiny PP}}}\xspace}

\newcommand{\pseudocode}{Algorithm}
\floatname{algorithm}{\pseudocode}

\iffalse

\else

\fi

\newcommand{\thetafull}{\ensuremath{\bm{\theta}_\text{full}}\xspace}
\newcommand{\thetafullmt}{\ensuremath{\thetafull^\text{MT}}\xspace}
\newcommand{\thetafullmthat}{\ensuremath{\hat{\bm{\theta}}_\text{full}^\text{MT}}\xspace}

\newcommand{\thetafullst}{\ensuremath{\thetafull^\text{ST}}\xspace}
\newcommand{\thetafullsthat}{\ensuremath{\hat{\bm{\theta}}_\text{full}^\text{ST}}\xspace}
\newcommand{\thetafullasrhat}{\ensuremath{\hat{\bm{\theta}}_\text{full}^\text{ASR}}\xspace}
\newcommand{\thetafullasr}{\ensuremath{{\bm{\theta}}_\text{full}^\text{ASR}}\xspace}

\newcommand{\pfull}{\ensuremath{p_{\text{full}}}}
\newcommand{\hatpfull}{\ensuremath{\hat{p}_{\text{full}}}}
\newcommand{\pwk}{\ensuremath{p_{\text{wait-$k$}}}}

\newcommand{\waitk}{\ensuremath{\text{wait-$k$}}\xspace}

\newcommand{\nextbeam}{\ensuremath{\mathit{next}\xspace}}


%% file: intro.tex

Simultaneous speech-to-text translation 
incrementally translates source-language speech into target-language text,
and is widely useful in many cross-lingual communication scenarios
such as international travels and multinational conferences.
The conventional approach to this problem is 
a cascaded one \cite{arivazhagan2020re,xiong2019dutongchuan,zheng2020fluent}, involving a pipeline of two steps.
First, the streaming automatic speech recognition (ASR) module
transcribes the input speech on the fly \cite{moritz2020streaming,wang2020low}, 
and then a simultaneous text-to-text translation module
translates the partial transcription into target-language text 
\cite{oda+:2014,dalvi:2018-incremental,ma+:2019,zheng2019simpler,zheng+:2019,zheng2020simultaneous,arivazhagan:2019}.


\begin{figure}[t!]
\centering
\includegraphics[width=1\linewidth]{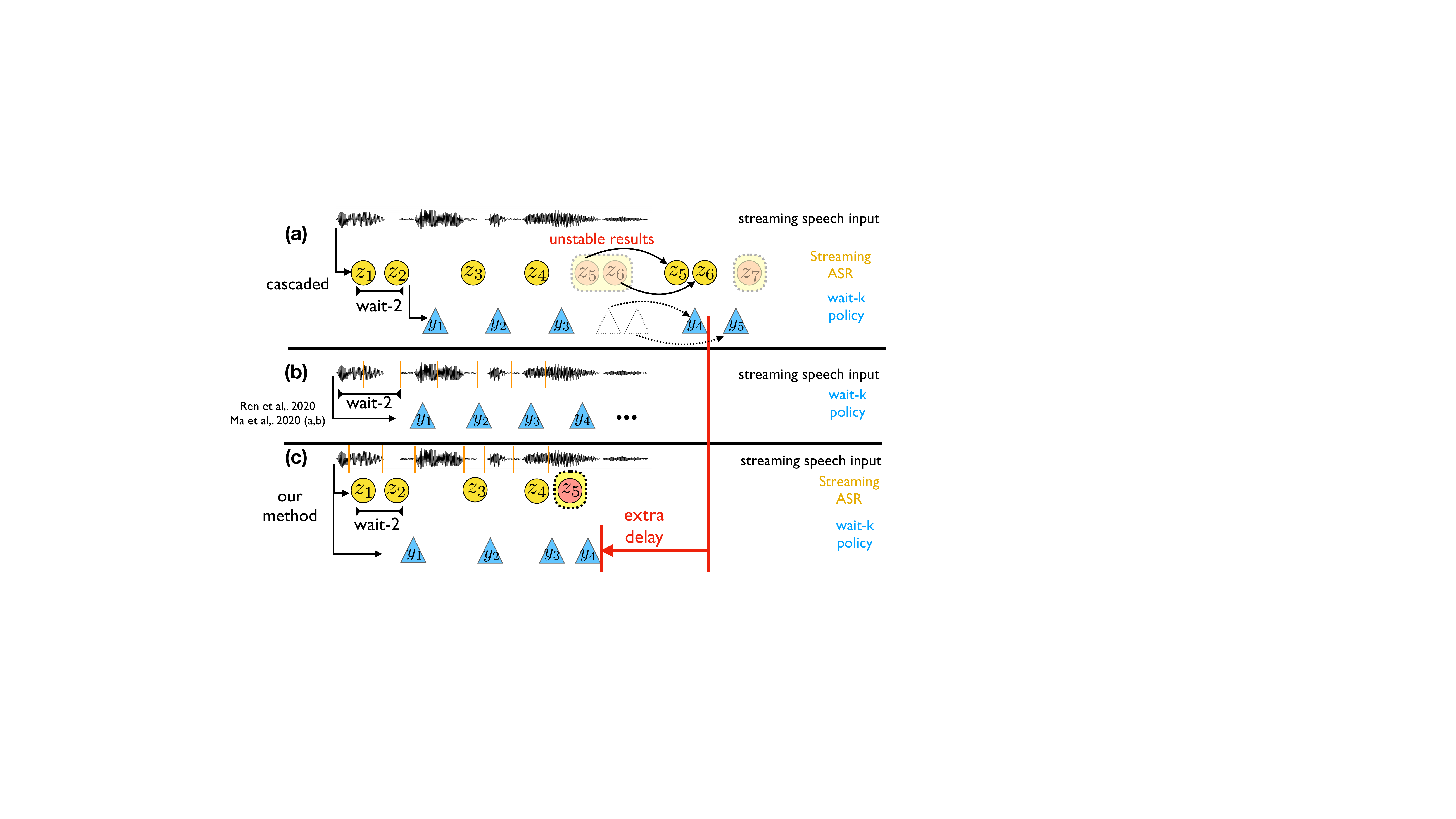}
\caption{Comparison between (a) cascaded pipeline, 
(b) direct simultaneous ST, and (c) our ASR-assisted simultaneous ST.
In (a), streaming ASR keeps revising some tail 
words for better accuracy, but causing extra delays to MT.
Method (b) directly translates  source speech without using ASR.
Our work (c)  
uses the intermediate results of the
streaming ASR module to  guide the decoding policy of (but not feed as input to) 
the speech translation module.
Extra delays
between ASR and MT are reduced in direct translation systems (b--c).}
\label{fig:comparison}
\end{figure}

However, the cascaded approach inevitably suffers from two limitations:
(a)  {\bf error propagation}, where
streaming ASR's mistakes confuse the translation module (which are trained on clean text), and
this problem worsens 
with noisy environments and accented speech;
and (b)  {\bf extra latency},
where the translation module has to wait until streaming ASR's output stabilizes, 
as ASR by default can repeatedly revise its output
(see Fig.~\ref{fig:comparison}).

To overcome the above issues, 
some recent efforts \cite{ren-2020-simulspeech,Ma2020StreamingSS,ma-2020-simulmt}
attempt to directly translate the source 
speech into target text simultaneously 
by adapting text-based \waitk strategy \cite{ma+:2019}. 
However, unlike simultaneous translation whose input is already segmented into words or subwords,
in speech translation,
the key challenge is to figure out the number of valid tokens within a given source
speech segment in order to apply the \waitk policy.
\citet{Ma2020StreamingSS,ma-2020-simulmt} simply assume a fixed number of words within a certain number of speech frames, which does not consider
various aspects of speech such as different speech rate, duration, pauses and silences,
all of which are common in realistic speech.
\namecite{ren-2020-simulspeech} design an extra Connectionist Temporal Classification (CTC)-based speech segmenter to detect 
the word boundaries in speech.
However, the CTC-based segmenter inherits the same shortcoming of CTC,
which only makes local predictions, thus limiting its segmentation accuracy.
On the other hand, to alleviate the error propagation,
\namecite{ren-2020-simulspeech} employ several different knowledge 
distillation techniques to learn the attentions of ASR and MT jointly.
These knowledge distillation techniques are complicated to train and it is an 
indirect solution for the error propagation problem.

We instead present a simple but effective solution (see Fig.~\ref{fig:decoding})
by employing two separate, but synchronized, decoders,
one for streaming ASR and the other for End-to-End Speech-to-text Translation (E2E-ST).
Our key idea is to use the intermediate results of streaming ASR 
to guide the decoding policy of, but not feed as input to, 
the E2E-ST decoder.
We look at the beam of streaming ASR 
to decide the number of tokens within the given source speech segment.
Then it is straightforward for the E2E-ST decoder to apply the \waitk policy and decide
whether to commit a target word or to wait for more speech frames.
During training time, we jointly train ASR and E2E-ST tasks
with a shared speech encoder in a multi-task learning (MTL) fashion 
to further improve the translation accuracy. 
We also note that having streaming ASR as an auxiliary output is extremely useful in real application scenarios where the user often wants to see both the transcription and the translation.
En-to-De and En-to-Es experiments on the MuST-C dataset demonstrate that our proposed technique achieves substantially better translation quality at similar level of latency.

%% file: prelim.tex
We formalize full-sentence tasks (ASR, MT and ST)
using the sequence-to-sequence framework,
and the streaming tasks (simultaneous MT and streaming ASR)
using the test-time \waitk method. 

\smallskip
\paragraph{Full-Sentence Tasks: ASR, NMT and ST}
The encoder first encodes the entire source input
into a sequence of hidden states;
in NMT, the input is a sequence of words, $\vecx=(x_1,x_2,...,x_m)$,
while in ASR and ST, we use $\vecs$ to denote the
input speech frames.
A decoder sequentially predicts target language 
tokens $\vecy=(y_1,y_2,...,y_n)$ in NMT and ST 
or transcription $\vecz$ in ASR,
conditioned on all encoder hidden states and previously committed tokens.
For example, the NMT model and its parameters \thetafullmt are defined as: 
\begin{align}
  \pfull(\vecy \mid \vecx; \thetafullmt) = \prod\nolimits_{t=1}^{|\vecy|}p(y_t \mid \vecx, \vecy_{<t}; \thetafullmt) 
   \notag\\
  \thetafullmthat = \argmax_{\thetafullmt} \prod\limits_{(\vecx,\vecy^*) \in D}\pfull(\vecy^* \mid \vecx;\thetafullmt) 
 \notag
\end{align}
Similarly,
we can obtain the
definitions for ASR ($\pfull(\vecz \mid \vecs; \thetafullasr)$)
and ST ($\pfull(\vecy \mid \vecs; \thetafullst)$).
Our model was learned from scratch in this work, 
but it can be improved with pre-training methods~\cite{zheng2021fused,chen2020mam}.

\paragraph{Simultaneous MT and Streaming ASR} 
In streaming decoding scenarios,
we have to predict target tokens
conditioned on the partial source input that is available.
For example,
the \textbf{test-time \waitk} method of \citet{ma+:2019} 
predicts each target token $y_t$ after reading source tokens $\vecx_{\leq t+k}$
using a full-sentence NMT model:
\begin{fleqn}
\begin{equation}
\hat{y}_t\! =\! \argmax_{y_t}\pwk(y_t \mid \vecx_{\leq t+k}, \hat{\vecy}_{<t};\thetafullmthat) 
\label{eq:pref2pref_decode}
\end{equation}
\end{fleqn}  
Intuitively speaking, \waitk only commits a new target word on receiving 
each new source word after an initial $k$ source words waiting.
Similarly, in the case of streaming ASR,
we could define $\hat{z}_t$ with growing 
speech chunks $\overline{\vecs}_i$ that are fed gradually.

\if
However, when we apply \waitk policy to direct 
simultaneous speech translation,
we need to detect the number of valid tokens within 
growing speech chunks first
to guide the decoding schedule.
A simple assumption of a certain fixed number of words with stable speech 
pace as described in \cite{Ma2020StreamingSS,ma-2020-simulmt} 
is inevitably being either conservative or aggressive for 
translation and leads to
either high latency or low accuracy.
\fi

%% file: methods.tex

In  text-to-text simultaneous translation,
the input stream is already segmented.
However,
when we deal with speech frames as source inputs,
it is not easy to determine the number of valid tokens within certain
speech segments.
Therefore, to better guide the translation policy, 
it is essential to detect the number of valid tokens accurately within low latency.
Different from the sophisticated design of speech segmenter in \namecite{ren-2020-simulspeech},
we propose a simple but effective method by using a synchronized streaming ASR and using its beam
to determine the number of words within certain speech segments.
Note that we only use streaming ASR for  source word counting,
but the translation decoder does not condition on any of  ASR's output.

\begin{figure}[t]
\centering
\includegraphics[width=1\linewidth]{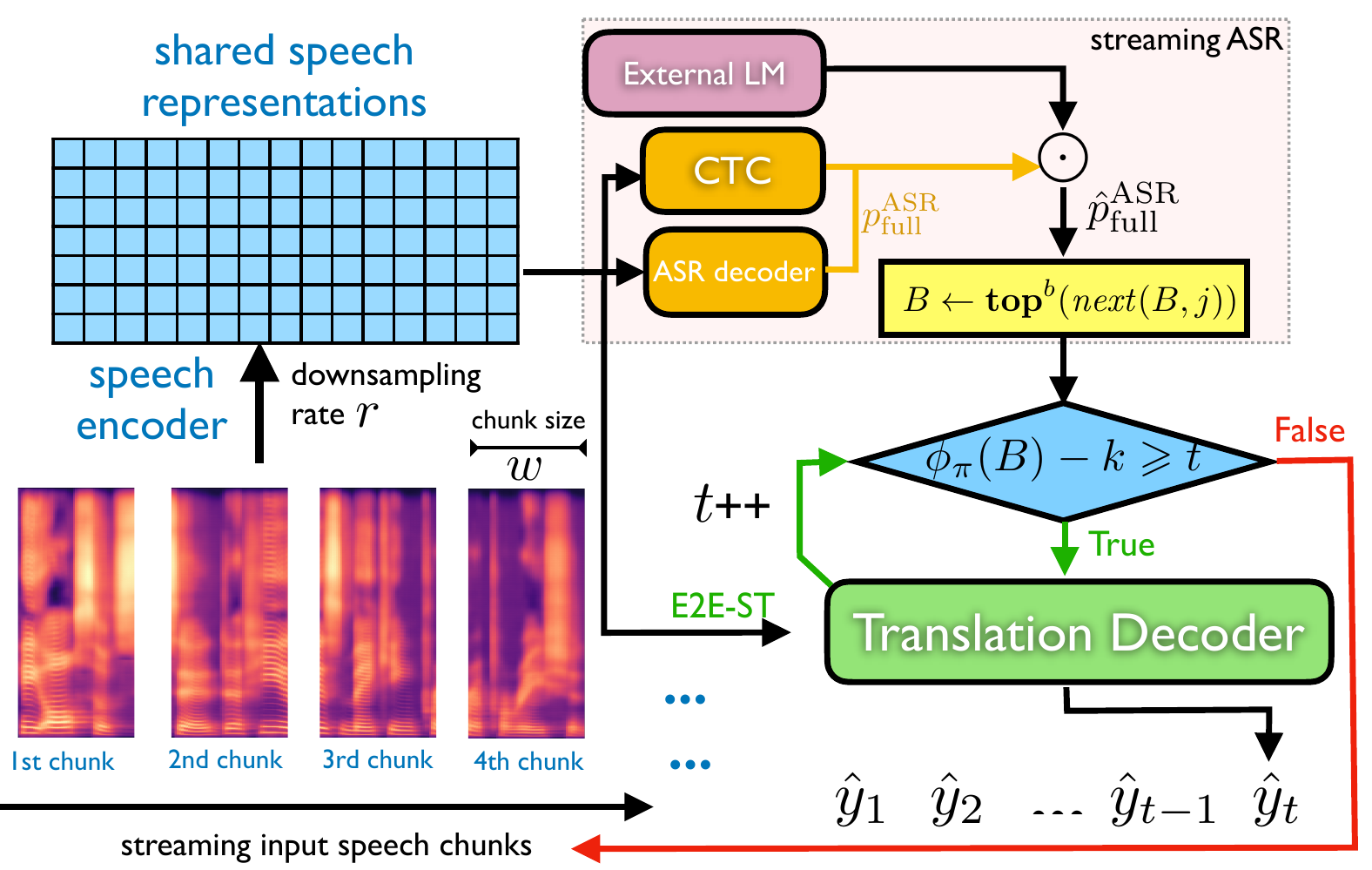}
\caption{Decoding  for synchronized streaming ASR and E2E-ST. 
Speech signals are fed into the encoder chunk by chunk. 
For each new-coming speech chunk, 
we look at the current streaming ASR beam  ($B$) to decide the
translation policy. See details in Algorithm~\ref{al:decode}.}
\label{fig:decoding}
\end{figure}

\subsection{Streaming ASR-Guided Simultaneous ST}

As shown in Fig.~\ref{fig:decoding},
at inference time, 
the speech signals are fed into the ST encoder 
by a series of fixed-size
chunks $\overline{\vecs}_{[1:i]}=[\overline{\vecs}_1,...,\overline{\vecs}_i]$,
where $w= |\overline{\vecs}_i|$ can be chosen from 
32, 48 and 64 frames of spectrogram.
\if
We use beam search, which are jointly scored by CTC,
an external language model and ASR decoder,
to search for the best decision of
the number of valid tokens at given time step.
\fi
As a result of the CNN encoder,
there is down sampling rate $r$ (e.g., we use $r=4$), 
from spectrogram to encoder hidden states.
For example, when we receive a chunk of  32 frames,
the encoder will generate 8 more hidden states.
In conventional streaming ASR,
the number of steps of  beam search is the same as the number of hidden states.

We denote $B_j$ to be the beam at time step $j$, which is an ordered list of 
size of $b$, and it expands to the next beam $B_{j+1}$ with the same size:
\begin{align*}
B_0 =& [\tuple{\startsym, \ \hatpfull^{\text{ASR}}(\startsym \mid \overline{\vecs}_0;\bm{\theta})}] \\
B_j =& \textstyle\toptop ^b  ( \nextbeam (B_{j-1}, j) )\\
\nextbeam (B, j)  = & 
     \{\tuple{\vecz \circ z_j, \ p \cdot \hatpfull^{\text{ASR}}(z_j \mid \overline{\vecs}_{\leq \tau(j)}, \vecz; \bm{\theta})} \mid\\
     &  \;\; \tuple{\vecz, p} \in B, z_j \in V \}
\end{align*}
where $\toptop^b(\cdot)$ returns the top $b$ candidates,
and $\nextbeam(B, j)$ expands the candidates from the previous step to the next step.
Each candidate is a pair $\tuple{\vecz, p}$,
where $\vecz$ is the current prefix and $p$ is
the accumulated probability from joint score between an external language model,
CTC and ASR probabilities, $\hatpfull^{\text{ASR}}$.
We denote the number of observable speech chunks at $j$ step as $\tau(j)=\ceil*{j*r/w}$.
And vice versa, for each new speech chunk, 
ASR beam search will advance for $w/r$ steps.

\begin{figure}[t]
\centering
\includegraphics[height=2.2cm]{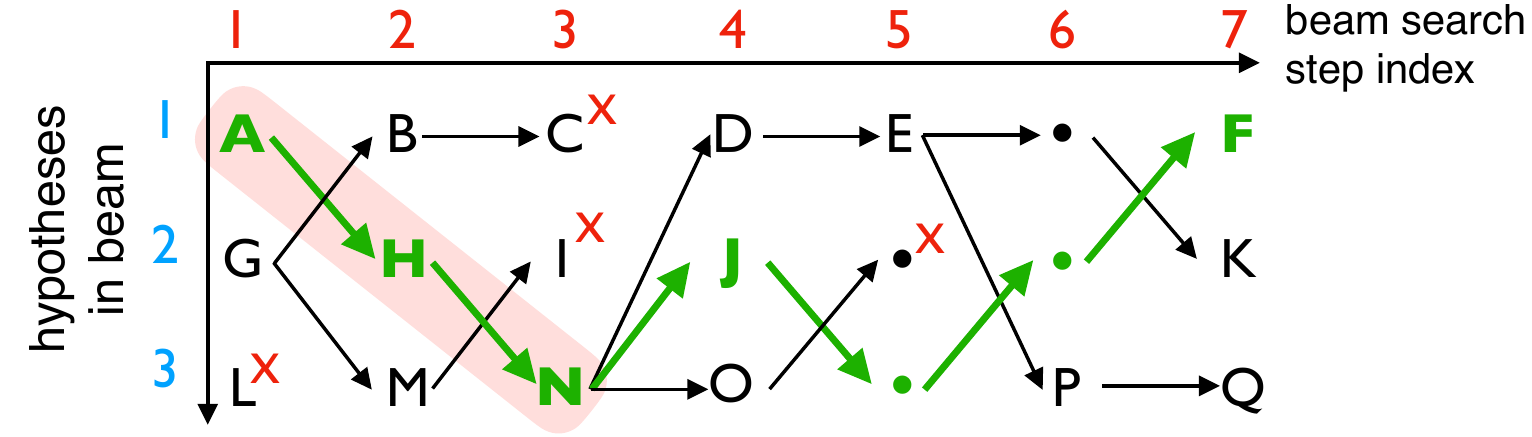}
\caption{An example of streaming ASR beam search with beam size 3. 
LCP is shaded in red ($\phi_\text{LCP}(B_7) \!=\! 3$); 
SH is highlighted in bold ($\phi_\text{SH}(B_7) \!=\! 5$). We use $\bullet$ to 
represent empty outputs in some steps
caused by CTC.}
\label{fig:beam}
\end{figure}

\begin{algorithm}
\caption{\footnotesize Streaming ASR-guided Simultaneous ST} 
\label{alg:rw}
\algrenewcommand\algorithmicindent{1em}
\algdef{S}[FOR]{REPEAT}[1]{\algorithmicrepeat\ {#1} }
\begin{algorithmic}[1]
\footnotesize
\State {\bf Input}: speech chunks $\overline{\vecs}_{[1:T]}$; $k$; $\phi_{\pi}(B_j)$; streaming decoding models: $\pfull^{\text{ST}}$ and $\hatpfull^{\text{ASR}}$
\State {\bf Initialize}: ASR and ST indices: $j = t = 0$; $B=B_0$
\For {$i = 1 \sim  T$ } \Comment{\textcolor{blue}{\footnotesize{feed speech chunks}}}
\REPEAT{$w/r$ steps} \Comment {\textcolor{blue}{  \footnotesize{do ASR beam search $w/r$ steps}}}
\State $B  \leftarrow  \toptop^ b( \nextbeam (B , j) )$; $j\text{++}$ \Comment{\textcolor{blue}{\footnotesize{ASR beam search}}}
\EndFor
\While {$\phi_{\pi}(B) -k \geqslant t$} \Comment{\textcolor{blue}{\footnotesize{new tokens?}}}
\State $\hat{y}_{t+1} \leftarrow  \pwk^{\text{ST}}(y_{t+1}  \mid \overline{\vecs}_{[1:i+1]}, \hat{\vecy}_{\le t};\thetafullst)$
\State \textbf{yield} \:$\hat{y}_{t+1}$; $t\text{++}$ \Comment{\textcolor{blue}{\footnotesize{commit translation to user}}}
\EndWhile
\EndFor
\end{algorithmic}
\label{al:decode}
\end{algorithm}


Note CTC often commits empty tokens $\epsilon$ due to empty speech frames,
and the lengths of different hypotheses within beam of streaming ASR are quite
different from each other. 
To take every hypothesis into consideration,
we design two policies to decide the number of valid tokens.

\begin{itemize}[leftmargin=10pt]
\setlength{\itemsep}{0pt}
\item \textbf{Longest Common Prefix} (LCP) uses the length of 
longest shared prefix in the  streaming ASR beam 
as the number of valid tokens within given speech.
This  is the most conservative strategy, which has  similar
latency to cascaded methods.

\item \textbf{Shortest Hypothesis} (SH) uses the length of 
shortest hypothesis in the current streaming ASR beam as the number of valid tokens. 
\end{itemize}

More formally, let $\phi_{\pi}(B)$  denote the number of valid
tokens in the beam $B$ under policy $\pi$:
\begin{eqnarray}
\phi_\text{LCP}(B) &=& \max \{ i \mid \exists \vecz', s.t. \forall \langle \vecz, c \rangle \! \in \! B, \vecz_{\leq i} \!=\! \vecz' \} \notag\\
\phi_\text{SH}(B) &=& \min \{ |\vecz| \mid \langle \vecz, c \rangle \in B \}  \notag
\end{eqnarray}
For example in Fig.~\ref{fig:beam}, 
$\phi_{\text{LCP}}(B_7) \!=\! 3$, $\phi_{\text{SH}}(B_7) \!=\! 5$. 
Also note that $\phi_\text{LCP} (B) \leq \phi_\text{SH}(B)$ for any beam $B$,
and that both policies are {\em monotonic}, i.e.
$\phi_\pi (B_j) \leq \phi_\pi (B_{j+1})$ for $\pi \in \{\text{LCP}, \text{SH}\}$ and all $j$.

Note we always feed the entire observable speech segments into
ST for translation,
and streaming ASR-generated transcription is not used for translation, 
so LCP might have similar latency with cascaded methods
but the translation accuracy is much better because more information 
on the source side is revealed to the translation decoder.

As shown in Algorithm \ref{al:decode},
during simultaneous ST, we  monitor the value of $\phi_{\pi}(B_j)$
while speech chunks are gradually fed into system.
When we have 
$\phi_{\pi}(B)-k \geqslant t$
where
$t$ is the number of translated tokens,
the ST decoder will be triggered to generate one new token as follows:
\begin{fleqn}
\begin{equation}
\hat{y}_t = \argmax_{y_t}\pwk(y_t \mid \overline{\vecs}_{[1:\tau(j)]}, \hat{\vecy}_{<t};\thetafullsthat)
\label{eq:st_decode}
\end{equation}
\end{fleqn}

\subsection{Joint Training between ST and ASR}

\begin{figure}[t]
\centering
\includegraphics[height=1.2cm]{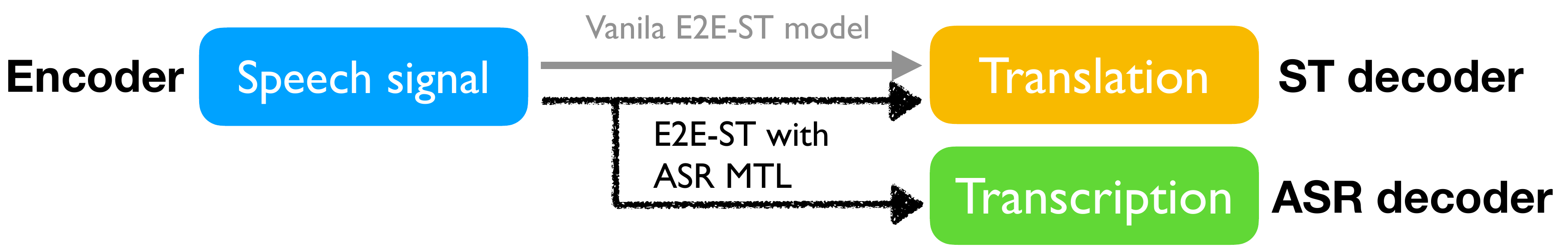}
\caption{We use full-sentence MTL framework to jointly learn ASR and ST
with a shared encoder.}
\label{fig:directmodel}
\end{figure}

Different from existing simultaneous translation solutions 
from~\cite{ren-2020-simulspeech,Ma2020StreamingSS,ma-2020-simulmt},
which make adaptations over vanilla E2E-ST architecture
as shown in  gray line of Fig.~\ref{fig:directmodel},
we instead use simple MTL architecture
which performs joint full-sentence training between ST and ASR:

\begin{align}
\thetafullsthat, \thetafullasrhat = \argmax_{\thetafullst, \thetafullasr} \!\!\!\prod\limits_{(\vecs,\vecy^*,\vecz^*) \in D} \!\!\!\! &\pfull^{\text{ST}}(\vecy^* \mid \vecs; \thetafullst) \notag\\
\cdot
&\pfull^{\text{ASR}}(\vecz^* \mid \vecs ; \thetafullasr)\notag
\end{align}

For ASR training, we  use  hybrid CTC/Attention
framework \cite{Watanabe2017}. 
Note that we train ASR and ST MTL with full-sentence fashion
for simplicity and training efficiency,
and only perform \waitk decoding policy at inference time.
Also, \thetafullst and \thetafullasr share the same speech encoder.


%% file: exp.tex

We conduct experiments on English-to-German (En$\to$De) and 
English-Spanish (En$\to$Es) translation on 
MuST-C~\cite{mustc}. 
We employ Transformer~\cite{vaswani+:2017} as the basic 
architecture 
and LSTM~\cite{hochreiter1997long} for LM.
For streaming ASR decoding we use a 
beam size of 5. 
Translation decoding is greedy due to incremental commitment.

Raw audios are processed with Kaldi~\cite{Povey11thekaldi} to extract 80-dimensional log-Mel filterbanks stacked
with 3-dimensional pitch features using a 10ms step size and a 25ms window size.
Text is processed by SentencePiece~\cite{kudo-richardson-2018-sentencepiece}
with a joint vocabulary size of 8K.
We take Transformer~\cite{vaswani+:2017} as our base architecture, 
which follows 2 layers of 2D convolution of size 3 with stride size of 2.
The Transformer model has 12 encoder layers and 6 decoder layers.
Each layer has 4 attention head with a size of 256.
Our streaming ASR decoding method follows~\citet{moritz2020streaming}.
We employ 10 frames look ahead for all experiments.
For LM, we use 2 layers stacked LSTM~\cite{hochreiter1997long} with 1024-dimensional hidden states, and set the embedding size as 1024.
LM are trained on English transcription from the corresponding language pair in MuST-C corpus.
For the cascaded model, we train ASR and MT models on Must-C dataset respectively, and they have the same Transformer architecture of our ST model.
Our experiments are run on 8 1080Ti GPUs.
And the we report the case-sensitive detokenized BLEU.

\begin{figure}[t]
    \centering
    \resizebox{\linewidth}{!}{\includegraphics{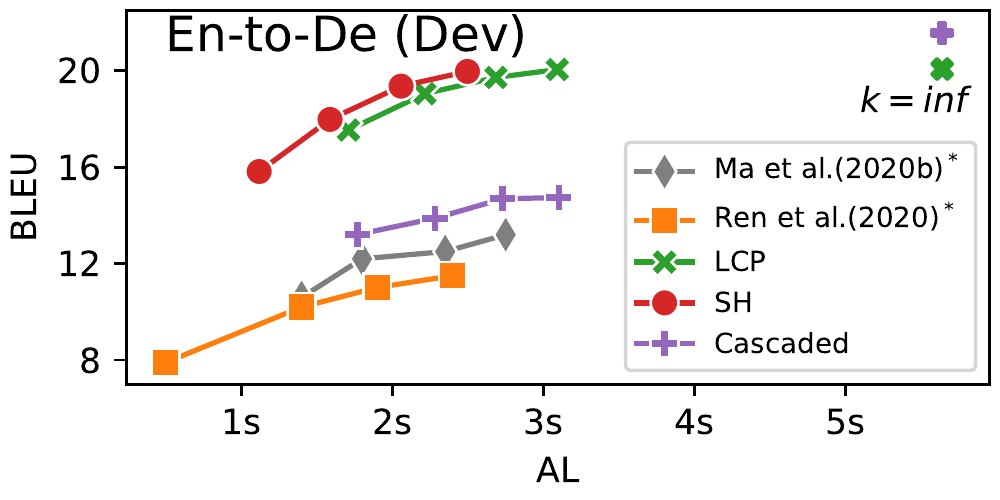}}\\
    \resizebox{\linewidth}{!}{\includegraphics{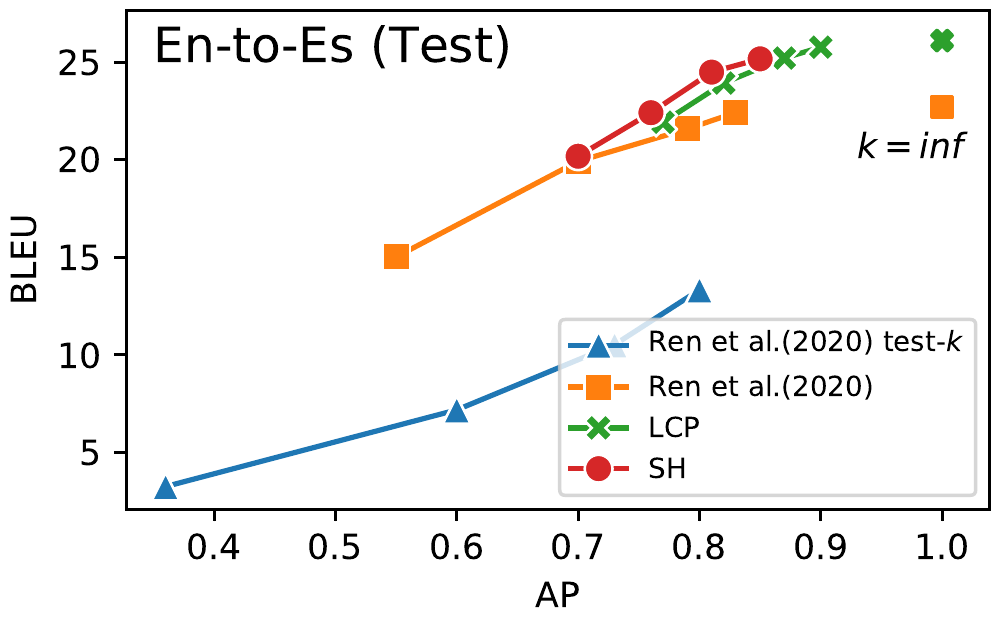}}
    \caption{Translation quality v.s. latency. 
    The dots on each curve represents different \waitk policy with 
    $k$=1,3,5,7 from left to right respectively.
    Baseline$^*$ results are from \citet{Ma2020StreamingSS}. 
    $k\!\!=\!\!\mathit{inf}$ is full-sentence decoding for  ASR and translation.
    test-$k$ denotes testing time \waitk.
    We use a chunk size of 48.}
    \label{fig:al}
    \end{figure}

\begin{figure*}[!h]\centering  \footnotesize

    \resizebox{\linewidth}{!}{
    \centering
    \setlength{\tabcolsep}{4pt}
    \renewcommand{\arraystretch}{0.5}
    \begin{tabu}{c | c  c  c  c  c  c  c}
    \toprule
    \rowfont{\small}
    chunk index  & 1 & 2 & 3 & 4 & 5 & 6 & end\\
    \midrule
    Gold transcript & can I be & \textbf{\textcolor{blue}{honest}} & \textit{SIL} & I don 't \textbf{\textcolor{orange}{love}} &  \textbf{\textcolor{orange}{that question}}  & \textit{SIL} \\
    Gold translation & \multicolumn{2}{c}{Darf ich \textbf{\textcolor{blue}{ehrlich}} sein ?} &  &  \multicolumn{2}{c}{Ich \textbf{\textcolor{orange}{mag diese Frage}} nicht .} &  \\
    \midrule 
    Streaming ASR & can I  & & & be \textbf{\textcolor{red}{on this}} I don 't  & \textbf{\textcolor{orange}{love that question}} & &  \\
    simul-MT wait-$3$  & & & & Kann ich \textbf{\textcolor{red}{da}} sein ? `` & Ich \textbf{\textcolor{orange}{liebe}} &  & \textbf{\textcolor{orange}{diese Frage}} nicht . \\
    \midrule
    SH wait-$3$ & & & & Kann ich \textbf{\textcolor{blue}{ehrlich}} sein ? Ich \textbf{\textcolor{orange}{liebe}} & \textbf{\textcolor{orange}{diese Frage}} & & nicht . \\
    LCP wait-$3$ & & & & Kann ich \textbf{\textcolor{blue}{ehrlich}} sein ? Ich  & & & \textbf{\textcolor{orange}{liebe diese Frage}} nicht . \\
    \bottomrule

    \end{tabu}
    }
   
  \caption{
  An example from the dev set of En$\to$De translation. 
  In the cascaded approach (streaming ASR + simul-MT wait-$3$),
  the ASR error (\textit{``on this''} for \textit{``honest''}) is propagated to the MT module,
  causing the wrong translation (\textit{``da''}).
  Our methods give accurate translations (\text{``ehrlich''}) with better latency (esp.~for the SH policy, the output of {\it ``diese Frage''} is synchronous with hearing {\it ``that question''}).
  \textit{``SIL''} denotes silence in speech.
  }
  \label{fig:case_study}
  \end{figure*}

\paragraph{Translation quality against latency}

\begin{figure}[t]
    \centering
    \includegraphics[width=0.95\linewidth]{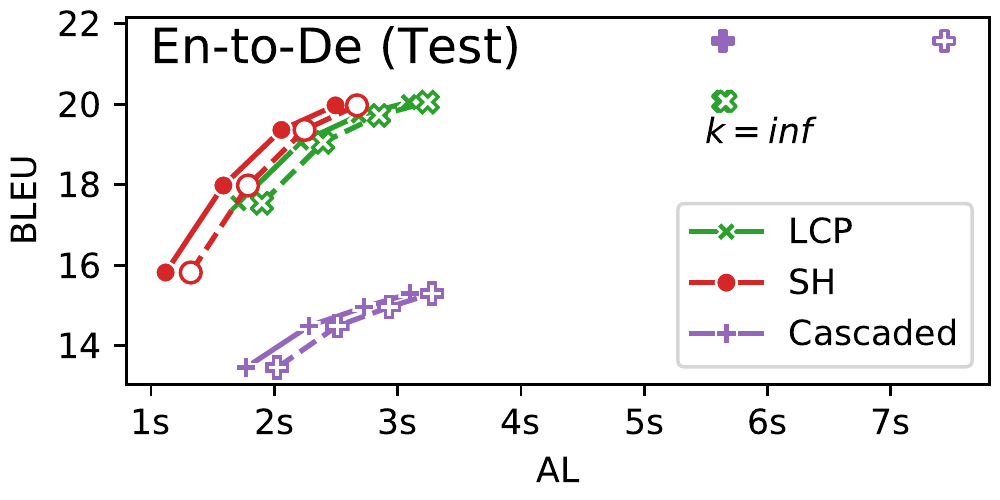}\\
    \includegraphics[width=0.95\linewidth]{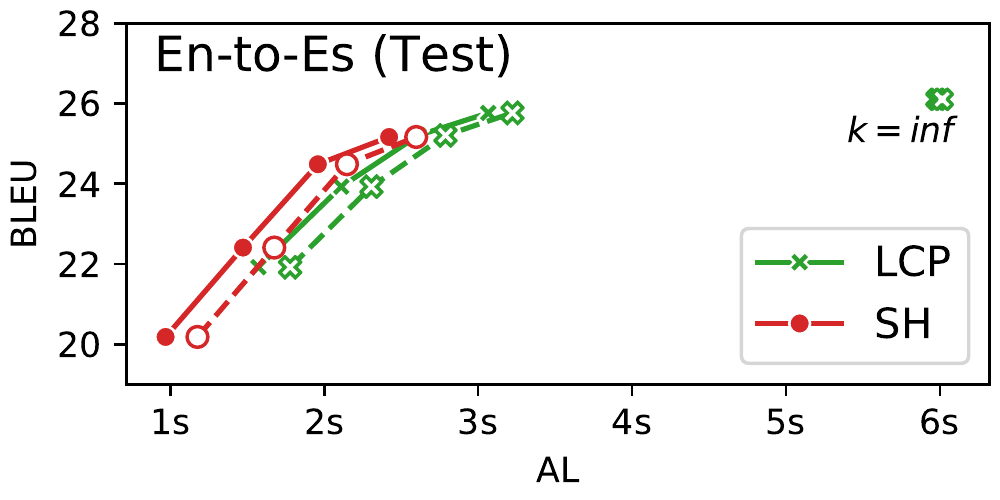}
    \caption{Translation quality against latency. 
    Each curve represents decoding with \waitk policy, 
    $k$=1,3,5,7 from left to right.
    The dashed lines and hollow markers indicate the latency considering the computational time.
    The chunk size is 48.
   }
    \label{fig:de_test_al}
\end{figure}

In order to clearly compare with related works, we evaluate the latency with AL defined in~\citet{Ma2020StreamingSS} and AP defined in~\citet{ren-2020-simulspeech}.
As shown in Fig.~\ref{fig:al},
for En$\to$De, results are on the dev set to be consistent with~\citet{Ma2020StreamingSS}. 
Compared with baseline models, our method achieves much better translation quality with similar latency.
To validate the effectiveness of our method,
we compare our method with~\citet{ren-2020-simulspeech} on En$\to$Es translation.
Their method does not evaluate the plausibility of the detected tokens, so it has a more aggressive decoding policy which results in lower latency.
However, our method  can still achieve better results with slightly lower latency.
Besides that, our model is trained in full-sentence mode, 
and only decodes with \waitk at inference time, which is very efficient to train.
Our test-time \waitk could achieve similar quality with their
genuine \waitk  (i.e., retrained) models which are very slow to train.
When we compare with their test-time \waitk, 
our model significantly outperforms theirs.

We further evaluate our method on the test set of En$\to$De and En$\to$Es translation.
As shown in Fig.~\ref{fig:de_test_al},    
compared with the cascaded model, our model has notable successes in latency and translation quality.
To verify the online usability of our model, 
we also show computational-aware latency.
Because our chunk window is 480ms, 
and the latency caused by the computation is smaller than this window size, 
which means that we can finish decoding the previous speech chunk when the next speech chunk needs to be processed, 
so our model can be effectively used online.

Fig.~\ref{fig:case_study} demonstrates that our method can effectively avoid the error propagation
and obtain better latency compared to the cascaded model.

\paragraph{Effect of chunk size and joint decision}
\begin{table}[t]\footnotesize
    \centering
    \resizebox{1\linewidth}{!}{
    \begin{tabular}{ccccccc}
    \toprule
    \multirow{2}{*}{Model} & \multicolumn{3}{c}{En$\to$De} & \multicolumn{3}{c}{En$\to$Es} \\
    \cmidrule(lr){2-4} \cmidrule(lr){5-7}
                                               & $w\!=\!32$       & $w\!=\!48$      & $w\!=\!64$       & $w\!=\!32$     & $w\!=\!48$ &  $w\!=\!64$ \\
    \midrule
    LCP                                      & 17.31      &   17.54         &  17.95         &  21.94  & 21.92 & 22.36   \\
    $-$ \textit{LM}                            & 14.60      &   15.66         &  15.91         &  18.54  & 19.15 & 19.95        \\
    $-$ \textit{LM}  \& \textit{AD}            & 13.76      &   14.82         &  15.26         &  17.42  & 18.06 & 19.32    \\
    \midrule
    SH                                       & 16.04        &   15.82        &  15.87        &  20.45  & 20.18 & 19.84    \\
    $-$ \textit{LM}                            & 13.76        &   14.01        &  13.84        &  17.31  & 17.21 & 17.78       \\
    $-$ \textit{LM}  \& \textit{AD}            & 10.44        &   11.25        &  11.65        &  13.61  & 14.27 & 14.62    \\
    \bottomrule
    \end{tabular}
    }
    \caption{BLEU score of wait-$1$ decoding with different chunk sizes and ASR scoring functions.
    \textit{AD} denotes ASR Decoder. \textit{LM} denotes Language Model.}
    \label{tb:ablation}
 \end{table}

Table~\ref{tb:ablation} shows that
the results 
are relatively stable
with various chunk sizes.
It can be flexible to balance the response frequency and computational ability.
We explore the effectiveness of ASR joint scoring,
and observe that
the translation quality drops a lot without LM.
Without LM and AD, 
our token recognition approach is similar to 
the speech segmentation in \citet{ren-2020-simulspeech},
which implies that their model is hard to 
segment the source speech accurately,
leading to unreliable translation decisions for ST.